\documentclass[10pt,twocolumn,letterpaper]{article}

\usepackage[pagenumbers]{cvpr} 

\usepackage{graphicx}
\usepackage{amsmath}
\usepackage{amssymb}
\usepackage{booktabs}

\usepackage{wrapfig}
\usepackage{bbm}
\usepackage{relsize}
\usepackage{empheq}
\newcommand{\cut}[1]{}
\usepackage{listings}
\usepackage{algorithm2e}
\usepackage{multirow}
\usepackage{makecell}
\usepackage{arydshln}
\usepackage{booktabs}
\usepackage{titling}
\usepackage{pifont}
\newcommand{\cmark}{\ding{51}}%
\newcommand{\xmark}{\ding{55}}%

\DeclareMathOperator*{\minimize}{minimize}

\usepackage[pagebackref,breaklinks,colorlinks]{hyperref}

\usepackage[capitalize]{cleveref}
\crefname{section}{Sec.}{Secs.}
\Crefname{section}{Section}{Sections}
\Crefname{table}{Table}{Tables}
\crefname{table}{Tab.}{Tabs.}

\def\cvprPaperID{6468} 
\def\confName{CVPR}
\def\confYear{2022}

\begin{document}
\RestyleAlgo{ruled}
\title{Partially Does It: Towards Scene-Level FG-SBIR with Partial Input}

\lstset{
    basicstyle=\ttfamily,
    keywordstyle=\color{blue},
    stringstyle=\color{DarkMagenta},
    commentstyle=\color{DarkGreen},
    morecomment=[l]{\%}
}

\author{Pinaki Nath Chowdhury\textsuperscript{1,2} \hspace{.2cm} Ayan Kumar Bhunia\textsuperscript{1} \hspace{.2cm} Viswanatha Reddy Gajjala\thanks{Interned with SketchX} \\ \hspace{.2cm} Aneeshan Sain\textsuperscript{1,2} \hspace{.2cm} Tao Xiang\textsuperscript{1,2} \hspace{.2cm} Yi-Zhe Song\textsuperscript{1,2} \\
\textsuperscript{1}SketchX, CVSSP, University of Surrey, United Kingdom.  \\
\textsuperscript{2}iFlyTek-Surrey Joint Research Centre on Artificial Intelligence.\\
{\tt\small \{p.chowdhury, a.bhunia, a.sain, t.xiang, y.song\}@surrey.ac.uk}
\vspace{-0.5cm}
}
\date{}
\maketitle

\begin{abstract}
    We scrutinise an important observation plaguing scene-level sketch research -- that a significant portion of scene sketches are ``partial". A quick pilot study reveals: (i) a scene sketch does not necessarily contain all objects in the corresponding photo, due to the subjective holistic interpretation of scenes, (ii) there exists significant empty (white) regions as a result of object-level abstraction, and as a result, (iii) existing scene-level fine-grained sketch-based image retrieval methods collapse as scene sketches become more partial. To solve this ``partial" problem, we advocate for a  simple set-based approach using optimal transport (OT) to model cross-modal region associativity in a partially-aware fashion. Importantly, we improve upon OT to further account for holistic partialness by comparing intra-modal adjacency matrices. Our proposed method is not only robust to partial scene-sketches but also yields state-of-the-art performance on existing datasets.

\end{abstract}
\vspace{-0.5cm}
\section{Introduction}
\label{sec:intro}
\begin{figure}
    \centering
    \includegraphics[width=\linewidth]{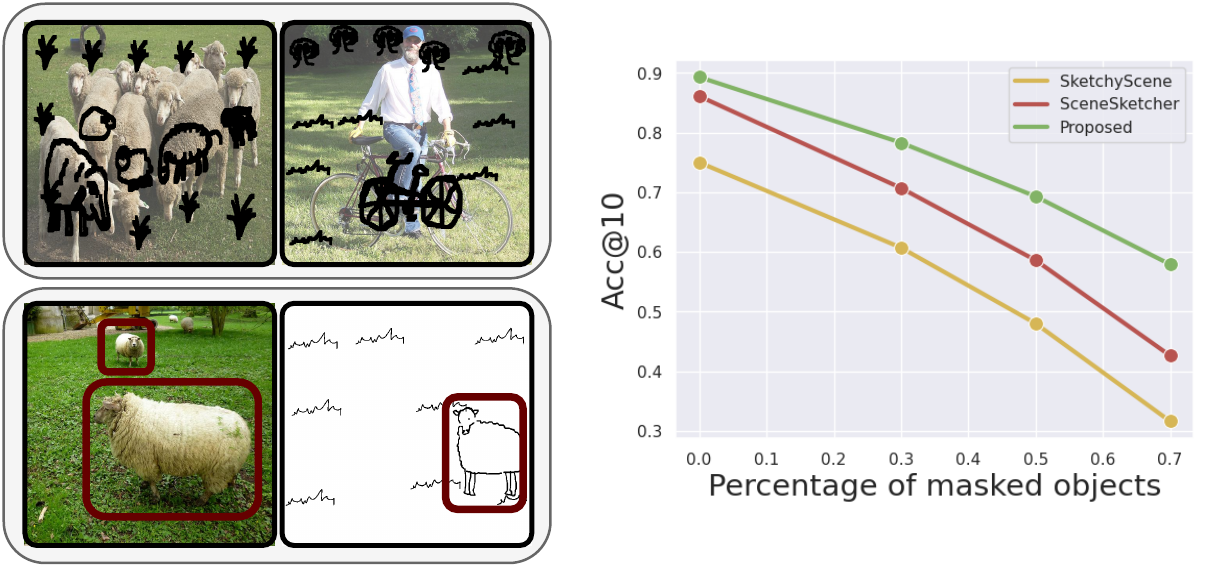}\\
    \vspace{-1.68in}
    \text{\hspace{1.49in} \textbf{(a)} \hspace{1.47in} \textbf{(b)}}\\[1.2in]
    \text{\hspace{-0.09in} \textbf{(c)}}
    \caption{(a): Scene sketches exhibit abstraction on global scene configuration as shown by overlapping sketches on top of their corresponding photos. (b): Existing scene-level FG-SBIR methods collapse as scene sketches become more partial. (c): There are significant empty (white) regions. Also, the sketch of a sheep in the scene might correspond to that in the centre of photo. This calls for a solution modelling region-wise associativity.}
    \label{fig:motivation}
    \vspace{-0.7cm}
\end{figure}

The prevailing nature of touch-screen devices has triggered significant research progress on sketches \cite{livesketch, bhunia2020pixelor, li2018sketch-r2cnn, liu2020scenesketcher, gao2020sketchyCOCO}. The field has predominately focused on object-level sketches to date \cite{tu-berlin, sketchy, ge2021creative}, studying its abstractness \cite{sketch-abstraction}, creativeness \cite{ge2021creative}, and applications such as image retrieval \cite{yu2016shoe, on-the-fly}, and 3D synthesis/editing \cite{lift3d}.
It was not until very recently that research efforts has undertaken a shift to scene-level analysis \cite{liu2020scenesketcher, sketchyscene, photosketching}. Compared to object-level sketches \cite{tu-berlin, sketchy}, scene sketches exhibit abstraction not only on individual objects, but also on global scene configurations. Fig. \ref{fig:motivation}(a) offers some examples where randomly  selected sketches are overlapped on top of their corresponding photos. 

We start with an important observation plaguing scene-level sketch research -- that a significant portion of scene sketches are ``partial". This partialness happens on two fronts (i) a scene sketch does not necessarily contain all objects in the corresponding photo, due to subjective interpretation of scenes, i.e., holistically partial, and (ii) there exists significant empty (white) regions as a result of object-level abstraction, i.e., locally partial. This is verifiable through a quick pilot study on existing scene sketch dataset \cite{gao2020sketchyCOCO}: (i) scene sketches include an average of $49.7\%$ of the objects present in photos. (ii) On average, only $13.0\%$ of the area in any scene sketch is occupied by its individual objects (the other $87.0\%$ being empty regions).

In this paper, we specifically tackle this ``partial" problem in the context of scene-level fine-grained sketch-based image retrieval (Scene-Level FG-SBIR). We first confirm the prevalence of this problem in existing Scene-Level FG-SBIR models, where we conduct an empirical study by measuring the retrieval accuracy by progressively masking out individual objects in a scene sketch. Fig.\ref{fig:motivation}(b) shows that popular models \cite{gao2020sketchyCOCO, sketchyscene} collapse as scene sketches become more partial, calling for a solution that is robust towards partial input, and in turn yields state-of-the-art performance. 

Global Average Pooling used by most existing FG-SBIR methods \cite{sketchSelf, sketchSemi} is clearly not up for the job since it is notorious for losing spatial scene configuration information. A naive alternative is computing distances between pairs of local features from corresponding regions in sketch and photo. This however is sub-optimal since sketch and photo do not follow strict region-wise alignment (see Fig. \ref{fig:motivation}). Alternatively, one can compute soft attentions independently in each domain \cite{deep-spatial-semantic}, yet this largely ignores the cross-modal gap between sketch and photo. Using cross-modal co-attention \cite{camp,sainBMVC} sounds a viable option but is otherwise intractable for practical applications\footnote{See Appendix for further discussion}. A close contender to our approach is using graph-based matching \cite{GOT, graph-cm-OT} of sketch and photo regions such as Liu \etal \cite{liu2020scenesketcher}. However, graph-based approaches have two common problems: (a) they dictate bounding-box annotations which are not always available (e.g., on \cite{sketchyscene}). (b) optimal graph construction (or sketching) strategy can be overly complicated \cite{graph-cm-OT, copt}.



The key to solving this ``partial" problem lies with modelling cross-modal region associativity. Crucially, such associativity needs to happen in a partially-aware fashion. This is because most sketch regions being empty will not be matched to any part of the photo. Partial graph matching is possible \cite{fused-gromov-wassertein, copt} but again not without resorting to expensive bounding-box annotations and complex scene graph construction procedures. Instead, as our first contribution, we advocate for the use of classic transportation theory (e.g., optimal transport (OT) \cite{boyd-convex}) to model this region associativity. Set-based approaches are a great fit because (i) they do not require any explicit data annotation, and (ii) naturally tackle this partial matching problem \cite{deepemd,graph-cm-OT}\footnote{Please refer to \cite{graph-cm-OT} for a detailed proof.}. 


While using OT can already model region associativity, it does not yet account for holistic partialness, i.e., differences in scene configurations. This dictates a holistic mechanism that accounts for spatial object relationships within each modal. Thus, as our second contribution, we improve upon OT by capturing intra-modal scene configurations for either modality in their respective region adjacency matrix \cite{spatialCorr-translation}. It follows that during cross-modal comparison, we compute the differences between these two matrices and obtain a scalar for each corresponding region to use alongside OT. Simply dotting together the intra-modal adjacency matrices is however not ideal since it ignores the local partialness of sketches which results in lots of near-zero entries in the adjacency matrix, ultimately leading to an overly sparse cross-modal matrix when dotted. Instead, we perform a weighted comparison by computing the cosine distance of two region-pairs, each pair taken from sketch and photo modality respectively.

In summary, our contributions are: (i) We show a significant portion of scene-level sketches are ``partial", both \emph{holistically} and \emph{locally}. (ii) The prevalence of this problem in existing FG-SBIR models is confirmed by showing how popular models \cite{sketchyscene, liu2020scenesketcher} collapse as scene sketches become more partial. (iii) We propose a simple solution to this ``partial" problem by modelling partially-aware cross-modal region associativity using the classic transportation problem (optimal transport (OT) \cite{boyd-convex}). (iv) We improve upon OT by capturing one-to-one intra-modal spatial relationships for partial scene sketches. (v) Our method is not only robust to partial input scene sketches but also yields state-of-the-art performance on existing scene sketch datasets.


\vspace{-0.2cm}
\section{Related Work}
\label{sec:related}
\vspace{-0.25cm}
\noindent \textbf{Fine-grained sketch based image retrieval} 
The ability of sketches to offer inherently fine-grained visual descriptions commences the avenues of fine-grained sketch-based image retrieval (FG-SBIR) \cite{sketch-crossdomain, generalisingFGSBIR, sketchCycle, sketchSemi, Sketch3T, DoodleIncremental, strokesubset}. This meticulous task aims to learn a pair-wise correspondence, for instance-level sketch-photo matching. Starting with graph-matching of deformable-part models \cite{DPM}, several deep learning approaches have surfaced with the advent of FG-SBIR datasets \cite{deep-spatial-semantic, text-sketch}. Yu \etal \cite{yu2016shoe} proposed a deep triplet-ranking model for instance-level matching. This framework was subsequently enhanced through a hybrid generative-discriminative cross-domain image generation \cite{sketch-crossdomain}, providing an attention based mechanism with advanced higher order retrieval loss \cite{deep-spatial-semantic}, utilising textual tags \cite{text-sketch}, or pre-training strategy \cite{sketchJigsaw, sketchSelf}. Unlike existing frameworks in FG-SBIR \cite{yu2016shoe, text-sketch} that independently map sketch and photo to a joint embedding sketch-photo space, Sain \etal \cite{sainBMVC} introduced a cross-modal co-attention mechanism for FG-SBIR to give considerable improvements in retrieval accuracy. Despite offering unmatched retrieval performance, it is often inapplicable in practice for large-scale retrieval, given the fact that every gallery photo needs to be compared with the query sketch every time for new retrieval. In this work, we push for a novel distance-metric function that would work at the output of independent sketch/photo branch, and models the region-wise associativity without any costly \cite{sainBMVC} pair-wise feature matching at the intermediate convolutional feature-map level. Our retrieval framework can be thought of as \emph{``best of both worlds"} i.e., \emph{fast-} \cite{yu2016shoe} and \emph{slow-} \cite{sainBMVC} retrieval \cite{thinking-fast-and-slow, retrieve-fast-rerank-smart}. In other words, we perform region-wise feature matching but still sketch/photo branch could be computed independently, unlike the necessity of repeated gallery image feature computation as in cross-modal co-attention mechanism \cite{sainBMVC}. 


\noindent \textbf{Scene-level sketches} As research in object-level (fine-grained) sketch-based image retrieval matured, recent works took the natural step towards the more practical but less explored setup of \emph{scene-}level for deeper and richer reasoning about sketched visual forms \cite{gao2020sketchyCOCO, sketchyscene, liu2020scenesketcher}. Zou \etal \cite{sketchyscene} studied segmentation and colorization on scene-level sketches. Gao \etal \cite{gao2020sketchyCOCO} proposed scene-sketch to photo generation  via generative adversarial approach using a sequential two-stage module. While Liu \etal \cite{liu2020scenesketcher} introduced FG-SBIR for scene-sketches using graph convolutional networks, it largely avoided the challenging setup of partial sketches by filtering existing datasets \cite{gao2020sketchyCOCO} having too few foreground instances (i.e., partial sketches). Unlike \cite{liu2020scenesketcher}, we propose a scene-level FG-SBIR setup which is robust to the more realistic setup of \emph{partial} sketches that lacks aligned, position-wise correspondence between instance-level sketch-photo pairs. 


\noindent \textbf{Dealing with Partial Data} One of the prolific areas studying incomplete or partial data is image inpainting \cite{pluralistic, wang2021inpainting, transfill}, where the objective is to generate (or fill) the missing (or masked) region by conditioning on the overall region. In the context of sketches, there are two broad lines of work: (a) two-stage pipeline that first tries to complete the partial sketch \cite{ha2018quickdraw, sketchHealer} by modelling a conditional distribution based on image-to-image translation followed by  performing task specific objective such as recognition \cite{sketchgan} or sketch-to-image \cite{interactiveSketch} generation. (b) single step framework \cite{on-the-fly} directly handle incomplete sketches to perform task specific objectives in a single step. {Similar to Bhunia \etal \cite{on-the-fly}, ours is a single step framework capable of handling incomplete or partial sketches. While existing literature \cite{on-the-fly, sketchgan} investigates object-level partial sketches, we focus on the novel setup of scene-level retrieval}.

\noindent \textbf{Application of Optimal Transport} The ability to learn structural similarity \emph{without explicit alignment information} makes optimal transport \cite{optimal-transport} in linear optimisation an important tool for several downstream tasks \cite{tensor-sift, pix2point, selfEMD, optimal-tracking, emd-visual-tracking}. Rubner \etal \cite{emd-retrieval} employed earth mover's distance, having the formulation of transportation problem, as a metric for color and texture based image retrieval. Later works extended optimal transport to the deep learning {landscape} for applications such as document classification \cite{optimal-word-embd}, few-shot learning \cite{deepemd}, domain adaptation \cite{deepJDOT}, self-supervised learning \cite{selfEMD}, neural machine translation \cite{volt}, and comprehending scene using 3D point cloud from monocular data \cite{pix2point}. In this work, for the first time, we study the application of optimal transport to design a differentiable distance metric function to model region-wise associativity without any explicit alignment label, and train a cross-modal retrieval system end-to-end through triplet-ranking objective.

\noindent \textbf{Learning unsupervised region-wise correspondence} Matching same or similar structure/content from two or more {input data} is a fundamental task in various downstream applications \cite{matching-survey, deepACG} such as image stitching \cite{bonny2016stitching}, image fusion \cite{ma2020fusion}, co-segmentation \cite{rgbd-cosegmentation, co-segmentation3d}, image retrieval \cite{zhou2011retrieval}, object recognition \cite{wohlhart2015recognition} and tracking \cite{wu2015tracking}. Region-wise correspondence could be learned in a supervised \cite{lift, verdie2015tilde, zhang2017local}, self-supervised \cite{zhang2018texture, superpoint, self-supervised-temporal-correspondence}, or unsupervised manner \cite{lenc2016covariant, savinov2017quad-networks, ono2018LF-NET, georgakis20183dmatching, keynet}. Self-supervised and unsupervised methods train without any human annotations and only use geometric \cite{halimi2019unsupervised-correspondence} or semantic \cite{yang-semantic-ot} constraints.
In contrast to these works, to the best of our knowledge, we here first try to bring the power of unsupervised region-wise associativity for cross-modal setup in an end-to-end trainable framework.  

\vspace{-0.2cm}
\section{Proposed Methodology}
\label{sec: proposed}
\vspace{-0.1cm}

\begin{figure}
    \centering
    \includegraphics[width=\linewidth]{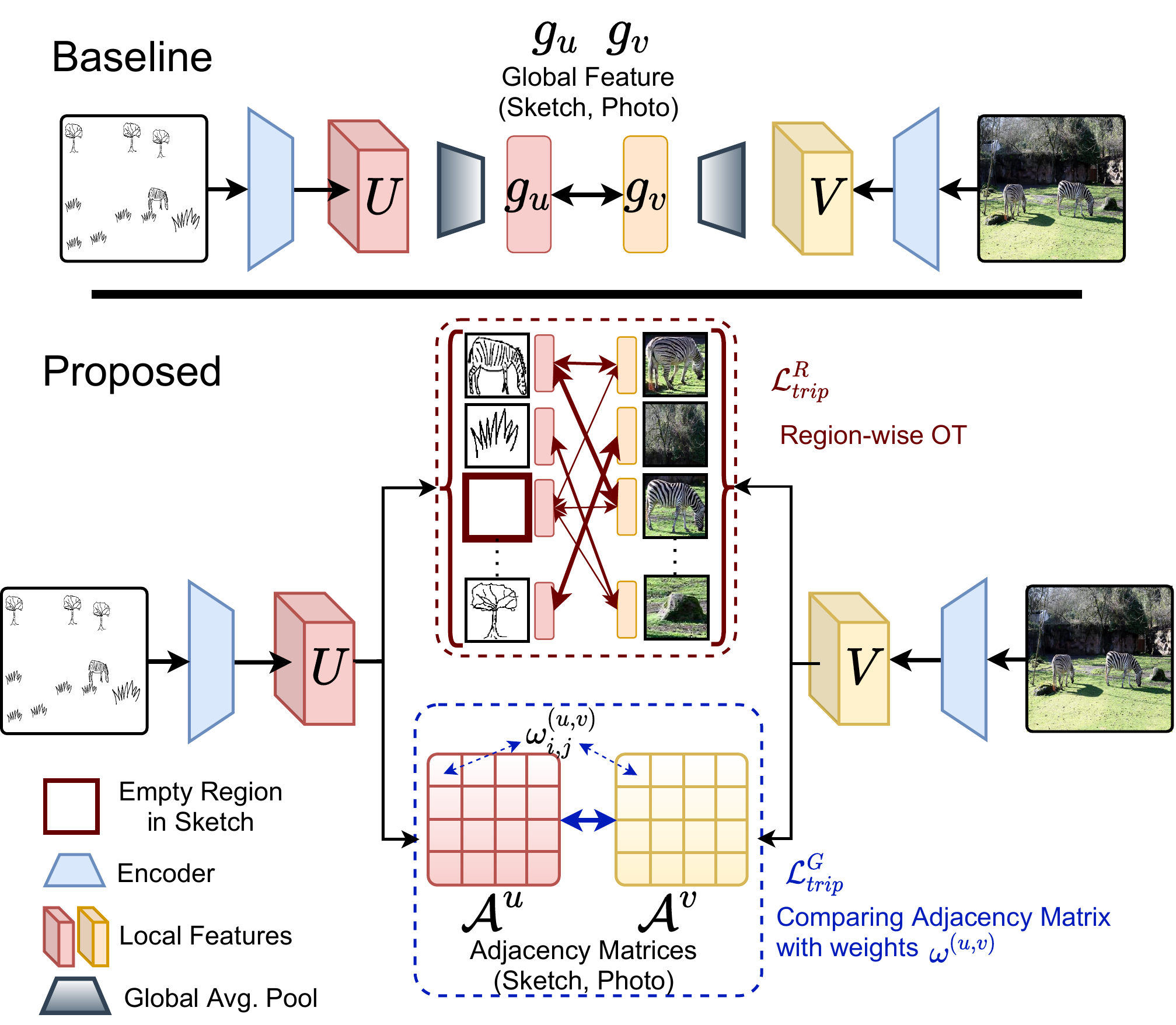}
    
    \caption{Illustration of our proposed method for scene-level FG-SBIR. Existing baselines typically use Global Average Pooling (GAP) on the convolutional feature-map. This loses localised region-specific feature representation necessary for ``partial" scene sketches. Our proposed method models partially-aware region-wise associativity to solve this ``partial" problem using: (i) set-based distance of local feature maps using optimal transport ($\mathcal{L}^{R}_{trip}$), (ii) a weighted cross-modal comparison of region adjacency matrix to capture holistic scene configuration ($\mathcal{L}^{G}_{trip}$).}
    \label{fig:proposed-method}
    \vspace{-0.5cm}
\end{figure}

Our objective is to retrieve the scene-image(s) which satisfy the partial sketch query constraint. The existing global feature vector-based representation, usually obtained through Global Average Pooling \cite{network-in-network},  ignores this \emph{partial associativity constraint}, thus failing to retrieve scene images from partial sketch faithfully. This work aims to model partial associativity by comparing corresponding region-wise localised features between every sketch-photo pair through a novel end-to-end trainable metric-learning loss so that the model can retrieve even from partial sketches. 

\vspace{-0.2cm}
\subsection{Baseline Retrieval Framework}
\label{sec: baseline-framework}
\vspace{-0.1cm}
First, we briefly summarise a baseline retrieval framework that remains state-of-the-art for SBIR literature to date. Given a sketch-photo pair represented as $(\mathcal{S}, \mathcal{I})$, a feature extractor $\mathcal{F}_{\theta}(\cdot)$ parameterised by $\theta $ is used to get the feature map $U = \mathcal{F}_{\theta}(\mathcal{S}) \ \in \mathbb{R}^{h_{S} \times w_{S} \times c}$ and $V = \mathcal{F}_{\theta}(\mathcal{I}) \in \mathbb{R}^{h_{I} \times w_{I} \times c}$ for sketch and photo, respectively. $\mathcal{F}_{\theta}$ can be modelled by CNN \cite{yu2016shoe}, LSTM \cite{li2018sketch-r2cnn}, Transformer \cite{lin2020sketchbert}, Graphs \cite{liu2020scenesketcher, sketchHealer, sketchlattics} or their combinations \cite{sketchSelf}. We flatten the backbone output feature-map $U \in \mathbb{R}^{h_{S} \times w_{S} \times c}$ and $V \in \mathbb{R}^{h_{I} \times w_{I} \times c}$ for sketch and photo as: $\mathbf{u} = \{{u_1}, \dots, {u_m}\}$ and $\mathbf{v} = \{{v_1}, \dots, {v_n}\}$, respectively, where $m=h_{s}w_{s}$,  $n=h_{I}w_{I}$, and $u_i, v_i\in \mathbb{R}^{c}$. Every vector of size $\mathbb{R}^{c}$ from either of $\mathbf{u}$ or $\mathbf{v}$ represents a localised regions specific feature. In order to get a single feature representation, we apply Global Average Pooling (GAP) and get ${g_{u}} = \frac{1}{m} \sum_{i=1}^{m} u_i \in \mathbb{R}^{c}$ and ${g_{v}} = \frac{1}{n}\sum_{j=1}^{n} v_j \in \mathbb{R}^{c}$ for sketch and photo, respectively. For training, the distance $d(\cdot,\cdot)$  to a sketch anchor $\mathcal{S}$ from a negative photo $\mathcal{I}^{-}$, denoted as $\beta^{-} = d({g_{u}}, {g_{v^{-}}})$
should increase while that from the positive photo 
$\mathcal{I}^{+}$, 
$\beta^{+} = d({g_{u}}, {g_{v^{+}}})$
should decrease. $d(a,b)$  can be either euclidean or cosine distance, but we consider dot product based cosine distance $(1 - a \cdot b)$ where $\{a, b\}$ are pre-normalised so that $||a||_{2}=1, \ ||b||_{2}=1$. Training is done via triplet loss with hyperparameter $\mu>0$: 
\vspace{-0.2cm}
\begin{equation}\label{eq: baseline-trip}
    \mathcal{L}_{trip} = \max\{ 0, \mu + \beta^{+} - \beta^{-}\}
\end{equation}


\noindent There are some inherent limitations of this standard baseline. \emph{Firstly}, applying GAP on the convolutional feature-map loses any localised region-specific feature representation, leaving no chance to learn region-wise associativity. 
\emph{Secondly}, region-wise associativity is a latent or hidden knowledge for which we do not have any explicit label, and this hidden knowledge is ignored here. \emph{Thirdly}, it assumes that every paired sketch is perfectly annotated containing \emph{all} the salient concepts/objects from the paired photo. However, in reality, most of the annotated sketches are partial and is biased towards the annotator's drawing skill and perception. Global feature vector-based representation will unnecessarily penalise mismatches for partial sketches. Therefore, this demands a further investigation on how to design a metric loss \cite{yu2016shoe, sainBMVC} that would implicitly discover the hidden region-wise associativity from partials sketches so that it generalises to query sketch with any degree of a partial instance during inference.

\subsection{Towards Partial Associativity}
 \label{sec:towards-partial}

\noindent \textbf{Reinterpretation of Baseline:}
The cosine distance between global sketch feature vector  $g_u \in \mathbb{R}^c$  and photo feature vector $g_v \in \mathbb{R}^c$ in our baseline could be reinterpreted as taking an average over all cosine distances computed between every localised region specific feature $u_i$ (sketch) and $v_j$ (photo).  Formally we can write $d({g_u}, {g_v})$ as follows,
\vspace{-0.3cm}
\begin{equation}\label{eq: analogy}
\resizebox{0.9\hsize}{!}{$%
\begin{split}
    d({g_u}, {g_v}) &= (1 - {g_{u}} \cdot {g_{v}}) = 1 - ( {\frac{1}{m} \sum_{i=1}^{m}} {u_i} ) \cdot ( {\frac{1}{n} \sum_{j=1}^{n}} {v_j} ) \\
    & \hspace{-1cm} = {\frac{1}{m \ n}\sum_{i=1}^{m} \sum_{j=1}^{n}} (1 - {u_i} \cdot {v_j}) = \frac{1}{m \ n} \sum_{i=1}^{m} \sum_{j=1}^{n} c_{i,j}
\end{split}$%
}
\end{equation}

\noindent \textbf{Weighted region-wise distance:}
This simple average operation gives equal weightage to every pair-wise distance  $c_{i,j}$ instead of prioritising those which actually have similar semantic meaning. Please see Fig. \ref{fig:proposed-method} for a visual illustration. Therefore, to model \emph{region-wise associativity} for measuring the distance between partial scene sketch-photo pairs, we extend the naive averaging of cosine distances (as in Eq. \ref{eq: analogy}) to a \emph{weighted region-wise cosine distance}:
\vspace{-0.15cm}
\begin{equation}
\begin{split}
    d_{W}(\mathbf{u}, \mathbf{v}) = \frac{1}{m \ n}\sum_{i=1}^{m} \sum_{j=1}^{n} c_{i,j} x_{i, j}
\end{split}
\end{equation}
We aim to learn the weights $x_{i,j}$ representing associativity between each pair of the localised feature from $\mathbf{u}$ (sketch) and $\mathbf{v}$ (photo) feature set, respectively. 
In other words, we compute all the pair-wise distances but give more weightage to those having similar semantics. We do not have any explicit labels for ${\mathcal{X}} \in \mathbb{R}^{m \times n}$, and to model this latent knowledge, we take inspiration from Optimal Transport \cite{convex-diff} literature.  Here, $x_{i,j}$ is termed as ``flow" from the source sketch region $u_i$ to destination photo region $v_j$. The task of finding the optimal flow $x_{i,j}$ for the given $c_{i,j}$ is similar to the classic transportation problem (TP) \cite{boyd-convex}. Overall, our objective is to design a weighted region-wise distance metric for a partial sketch to photo matching.   

\noindent \textbf{Optimal Transport}:
In classical transportation problem, $m$ suppliers $\mathrm{S} = \{s_i | i=1, \dots, m\}$ are required to supply $n$ demanders $\mathrm{D} = \{d_j | j=1, \dots, n\}$. The cost of transportation $c_{i,j}$ for a unit goods between the $i^{th}$ supplier and $j^{th}$ demander is $c_{i, j} = (1 - {u_i} \cdot {v_j})$. The optimisation objective of TP is to find the least expensive \emph{``flow"} of goods from suppliers to demanders, represented by $\tilde{\mathcal{X}} \in \mathbb{R}^{m \times n}$. 
Optimising $\tilde{\mathcal{X}}$ is analogous to our aim of prioritising those region-wise distances which are semantically similar, thus modelling pair-wise associativity from two feature set $\mathbf{u}$ and $\mathbf{v}$. TP objective is written as:
\vspace{-0.2cm}
\begin{equation}\label{eq: OT}
\begin{split}
    & \minimize_{\mathcal{X}} \sum_{i=1}^{m} \sum_{j=1}^{n} c_{i,j}x_{i,j} \\
    \text{subject to } & x_{i,j} \geq 0, \ i={1, \dots, m}, \ j={1, \dots, n} \\
    &\hspace{-1.7cm} \sum_{j=1}^{n} x_{i,j} = s_i, \ \sum_{i=1}^{m} x_{i,j} = d_j; \ i=[1, m] \ j=[1, n]
\end{split}
\end{equation}
The total flow from the $i^{th}$ region in a query sketch to every region in the target photo is represented by $s_i$, calculated as: $s_i = \sum_{j=1}^{n} x_{i,j}$. Similarly, $d_j$ denotes the total flow from all $m$ regions in the query sketch to the $j^{th}$ region in the target photo, computed as $d_j = \sum_{i=1}^{m} x_{i,j}$. This optimisation problem (Eq. \ref{eq: OT}) falls in the category of linear programming \cite{boyd-convex} as the objective and constraints are all affine and can be solved using the classical interior-point method. A naive solution to TP is intractable \cite{QPTH} \cut{non-differentiable}. Therefore, to bring the power of optimal transport for modeling region-wise partial associativity in case of partial sketch-based scene retrieval, we need a differentiable solution \cut{differentiable approximation} for end-to-end training.  

\noindent \textbf{Differentiability of the Solution:}
Our objective is to make the flow $\mathcal{X}$ differentiable with respect to model parameters $\theta$. Towards that goal, we rewrite Eq. \ref{eq: OT} in a parametric convex optimisation \cite{convex-diff} form involving model parameter $\theta$ as:
\vspace{-0.3cm}
\begin{equation}\label{eq: OT-compact}
\begin{split}
    & \hspace{-1cm} \minimize_{\mathcal{X}} c(\theta)^{T} \mathcal{X} \\
    \hspace{-0.3cm} \text{subject to } & f(\mathcal{X}, \theta) \preceq 0; \hspace{0.2cm}  h(\mathcal{X}, \theta) = b (\theta)
\end{split}
\end{equation}
\noindent where, $f(\mathcal{X}, \theta)$ is equivalent to the inequality constraint $x_{i,j} \geq 0$, and $h(\mathcal{X}, \theta) = b(\theta)$ is the equality constraint equivalent to $\sum_{j=1}^{n}x_{i,j}=s_i$ and $\sum_{i=1}^{m}x_{i,j}=d_j$ for all $i=1, \dots, m$; $j=1, \dots, n$. 

\noindent In order to combine Eq.~\ref{eq: OT-compact}, involving three fragmented parts (one optimisation objective, one equality, and one inequality constraints) into a single differentiable equation, we augment the objective function with a weighted sum of its equality and inequality constraints, defined as Lagrangian:
\vspace{-0.3cm}
\begin{equation}\label{eq: lagrange}
\begin{split}
    \mathrm{L}(\mathcal{X}, \lambda, \nu, \theta) & = \  c(\theta)^{T}\mathcal{X} 
    \\ & + \lambda^{T}f(\mathcal{X}, \theta) + \nu^{T}(h(\mathcal{X}, \theta) - b(\theta))
\end{split}
\end{equation}
The vectors $\lambda \geq 0$ and $\nu$ are the dual variables or Lagrange multiplier vectors. From Eq. \ref{eq: lagrange} we are looking for the optimal value of $\{\tilde{\mathcal{X}}, \tilde{\lambda}, \tilde{\nu}\}$ which corresponds to the minimum possible scalar value returned by $\mathrm{L}(\mathcal{X}, \lambda, \nu, \theta)$. For Eq. \ref{eq: lagrange},  assuming Slater's condition holds, then the necessary and sufficient conditions for optimality of $\mathrm{L}(\mathcal{X}, \lambda, \nu, \theta)$ is given by the Karush-Kuhn-Tucker (KKT) conditions \cite{boyd-convex}, which can be algebraically represented as:
\begin{equation}\label{eq: KTT}
    g(\tilde{\mathcal{X}}, \tilde{\lambda}, \tilde{\nu}, \theta) = \left[ \begin{array}{c}
         \nabla_{\mathcal{X}} \mathrm{L}(\tilde{\mathcal{X}}, \tilde{\lambda}, \tilde{\nu}, \theta) \\
         \mathbf{diag} (\tilde{\lambda}) f(\tilde{\mathcal{X}}, \theta) \\
         h(\tilde{\mathcal{X}}, \tilde{\theta}) - b(\theta)
    \end{array} \right] = 0
\end{equation}
where $\mathbf{diag}(\cdot)$ transforms a vector into diagonal matrix. The Jacobian $\nabla_{\theta}(\tilde{\mathcal{X}})$, provides the differentiation of our region-wise associativity $\mathcal{X}$ with respect to model parameters $\theta$ to allow end-to-end training. It can be derived from Eq. \ref{eq: KTT} using Dini classical implicit function theorem \cite{implicit-function} as,
\vspace{-0.3cm}
\begin{equation}\label{eq: theorem}
    \nabla_{\theta} (\tilde{\mathcal{X}}) = -\nabla_{\mathcal{X}} g(\tilde{\mathcal{X}}, \theta)^{-1} \nabla_{\theta} g(\tilde{\mathcal{X}}, \theta)
\end{equation}

\noindent \textbf{Determining the equality constraints $s_i$ and $d_j$:}
Earlier we formulated a differentiable way to calculate region-wise associativity $\mathcal{X}$, but we need to define the two important parameters  $s_i$ and $d_j$ of Eq. \ref{eq: OT}. Our assumption is that summation of all the region-wise associativity values should be equal to the cosine similarity between the global feature vector representation of given sketch ($g_u$) and photo ($g_v$). Note, while we constraint the total (\texttt{sum}) value of $\mathcal{X} \in \mathbb{R}^{m \times n}$, the model is free to decide 
how to distribute the total value to individual region-wise associativity ($x_{i,j}$) to achieve optimality. Given Global Average Pooled vector $g_u = \frac{1}{m} \sum_{i=1}^{m} u_i$ and $g_v = \frac{1}{n} \sum_{j=1}^{n} v_j$ from sketch and photo, respectively, the summation over all $x_{i,j}$ (total value) can be formally written as:
\vspace{-0.3cm}
\begin{equation}\label{eq: OT-global}
\begin{split}
    \sum_{i=1}^{m}\sum_{j=1}^{n} x_{i,j} &= g_u \cdot g_v = ( \frac{1}{m} \sum_{i=1}^{m} {u_i} ) \cdot ( \frac{1}{n} \sum_{j=1}^{n} {v_j}) \\
    & \hspace{-1.5cm} = \frac{1}{m\ n} \sum_{i=1}^{m} ( \sum_{j=1}^{n} {u_i} \cdot {v_j} ) \text{ (distributive property)}
\end{split}
\end{equation}
Therefore, following equality constraints in Eq. \ref{eq: OT} and ignoring the constant $(\frac{1}{mn})$
\vspace{-0.3cm}
\begin{equation}\label{eq: OT-equality}
    \sum_{j=1}^{n}x_{i,j} = \sum_{j=1}^{n} {u_i} \cdot {v_j} \implies
    s_i = {u_i} \cdot \sum_{i=1}^{n} {v_j}
\end{equation}
\text{similarly, } $d_{j}={v_j} \cdot \sum_{i=1}^{m} {u_i}$ Hence, our modified distance-metric function $d_{W}(\cdot, \cdot)$, which measures the scalar distance between two feature sets, modelling region-wise associativity between sketch and photo, is computed as:
\vspace{-0.4cm}
\begin{equation}\label{eq: OT-weighted}
\begin{split}
    d_W (\mathbf{u}, \mathbf{v}) & = \frac{1}{m \ n} \sum_{i=1}^{m} \sum_{j=1}^{n} (1 - u_i \cdot v_j) {\tilde{x}_{i,j}} \\
    \mathcal{L}_{trip}^{R} & = \max\{0, \mu_w + \beta^{+}_{R} - \beta^{-}_{R}\}
\end{split}
\end{equation}
\vspace{-0.3cm}

where, $\beta^{+}_{R} = d_W(\mathbf{u}, \mathbf{v^{+}})$ is the distance to a sketch anchor $\mathcal{S}$ to positive photo $\mathcal{I}^{+}$. Similarly, we calculate $\beta^{-}_{R}$ for negative photo $\mathcal{I}^{-}$. $\mu_w$ is margin hyperparameter.
 
\subsection{Preserving scene structure consistency}
\label{sec:scene-structure}
While optimal transport helps to model localised region-wise associativity to measure the distance between sketch-photo pairs, it cannot preserve the global structural consistency \cite{fused-gromov-wassertein} necessary for fine-grained retrieval. Region-wise associativity may fail to distinguish between scene sketches where individual objects are similar, but the global spatial arrangement is different.  For instance, moving a sketched ``tree" from the top-left corner to the bottom right will result in a similar distance value. Therefore, we argue that while 
${d_W}(\cdot)$ matches the features at the local level it is sub-optimal at considering the global spatial information. 


We design the global structural consistency through \emph{adjacency matrix} formulation which  captures spatially-correlative maps to explicitly represent the global scene structure. Given sketch and photo feature set $\mathbf{u} = \{{u_1}, \dots, {u_m}\}$ and $\mathbf{v} = \{{v_1}, \dots, {v_n}\}$ respectively, we compute their respective adjacency matrix as:
\vspace{-0.2cm}
\begin{equation}\label{eq:adjacency}
\begin{split}
    \mathcal{A}^{u}_{i,j} &= \frac{1}{m \times m} \frac{u_i \cdot u_j}{||u_i||_{2} \ ||u_j||_{2}}; \ \mathcal{A}^{u} \in \mathbb{R}^{m \times m} \\
    \mathcal{A}^{v}_{i,j} &= \frac{1}{n \times n} \frac{v^{}_i \cdot v^{}_j}{||v^{}_i||_{2} \ ||v^{}_j||_{2}}; \ \mathcal{A}^{v} \in \mathbb{R}^{n \times n} \\
\end{split}
\end{equation}
Our adjacency matrix essentially computes region-wise self-similarity in sketch and photo modality. Naively comparing $(\mathcal{A}^{u}, \mathcal{A}^{v})$ assumes that \emph{all} regions within a particular modality (sketch or photo) contribute towards self-similarity \cite{spatialCorr-translation}. However, this assumption does not hold for partial scene sketches with empty (sparse) or uncorrelated regions. Hence, we introduce a weighting factor $\omega_{i,j}^{u,v}$ that provides lower importance while comparing $(\mathcal{A}^{u}_{i,j}, \mathcal{A}^{v}_{i,j})$ if there is a low correlation between either of $(u_i, v^{}_{i})$, $(u_i, v^{}_{j})$, $(u_j, v^{}_{i})$, or  $(u_j, v_{j})$.  The distance function capturing global structural consistency is as follows:
\vspace{-0.3cm}
\begin{equation}
\begin{split}\label{dg_eqn}
    d_{G} (\mathbf{u}, \mathbf{v}) = \sum_{i=1}^{m} \sum_{j=1}^{m} \omega^{(u,v)}_{i,j} || \mathcal{A}^{u}_{i,j} - \mathcal{A}^{v}_{i,j}||_{1} \\
    \omega^{(u,v)}_{i,j} = (u_i \cdot v^{}_i) (u_i \cdot v^{}_j) (u_j \cdot v^{}_i) (u_j \cdot v^{}_j)
\end{split}
\end{equation}

if $m \neq n$, we make use of bi-linear interpolation \cite{wide-narrow} to make $\mathcal{A}^{v}$ to be of same spatial size as $\mathcal{A}^{u}$ to realize matrix subtraction.  
Intuitively, if either of the localised partial sketch feature $(u_{i}, u_{j})$ be a sparse region, it will have a lower correlation with the localised photo feature set $(v_{i}, v_{j})$ that will give a low value of $\omega^{(u,v)}_{i,j}$ in the distance function $d_{G}(\cdot, \cdot)$. Such a mechanism ignores empty and uncorrelated regions in the partial sketch while focusing only on the relevant regions. Hence, we can define our final training objective as,
\vspace{-0.2cm}
\begin{equation}\label{eq:total}
\begin{split}
    \mathcal{L}_{total} &= \mathcal{L}_{trip}^R + \alpha \cdot \mathcal{L}_{trip}^G \\
    \mathcal{L}_{trip}^G &= \max\{0, \mu_g + \beta^{+}_{G} - \beta^{-}_{G}\} \\
\end{split}
\end{equation}
where, $\beta^{+}_{G} = d_G(\mathcal{A}^{u} \mathcal{A}^{v^{+}})$, computed by Eqn. \ref{dg_eqn}. Similarly we compute $\beta^{-}_{G}$. $\alpha, \mu_g$ are hyperparameters.

\section{Experiments}
\noindent \textbf{Datasets:} We use two benchmark scene sketch datasets that support scene-level FG-SBIR tasks: (a) SketchyScene \cite{sketchyscene} comprise of sketch templates with paired cartoon style photos. Following Zou \etal \cite{sketchyscene}, we adopt the standard train/test split of $2472/252$ of sketch-photo pairs for scene-level FG-SBIR. On average each scene sketch has $16$ instances, $6$ object classes, and $7$ occluded instances. (b) Unlike SketchyScene \cite{sketchyscene} that comprise of cartoon style photos, SketchyCOCO \cite{gao2020sketchyCOCO} includes natural photos from COCO Stuff dataset \cite{caesar2018cocostuff} with paired scene sketches. Following Liu \etal \cite{liu2020scenesketcher}, we use $1015/210$ train/test split of sketch/photo pairs. In addition, we also evaluate how our proposed method generalise to object-level sketches in QMUL-Shoe-V2 \cite{yu2016shoe}. QMUL-Shoe-V2 contains $6,730$ sketches and $2,000$ photos. Following \cite{on-the-fly} we use $6,051$ and $1,800$ respectively for training and the rest for testing.

\noindent \textbf{Implementation Details:} Our model is implemented in PyTorch on a 11GB Nvidia RTX 2080-Ti GPU. ImageNet pretrained InceptionV3 \cite{inceptionV3} (excluding auxiliary branch) is used as the encoder network $\mathcal{F}_{\theta}(\cdot)$ where we remove the global average pooling and flattening layers. We train our model for $200$ epochs using Adam optimiser with learning rate $0.0001$, batch size $16$, margin value for triplet loss as $0.3$, and value of $\alpha$ in Eq. \ref{eq:total} is set to $0.01$. Our novel distance metrics for FG-SBIR do not add training parameters, hence the total number of model parameters stays the same as its backbone network. During training, to solve the linear programming problem in Eq. \ref{eq: KTT} we use a GPU accelerated \emph{convex optimisation solver} QPTH proposed in Amos and Kolter \cite{QPTH} to derive gradients for backpropagation. Since gradients only needs to be computed during training, for testing we replace the QPTH solver with a faster alternative in the OpenCV library \cite{opencv} that is non-differentiable\footnote{The differentiable QPTH solver in Amos and Kolter \cite{QPTH} use a custom primal-dual interior point method which is slower compared to the non-differentiable solver in OpenCV employing a modified Simplex algorithm.}.
 A PyTorch-like pseudo-code to solve the linear programming problem using QPTH and to compute region-wise associativity $x_{i,j}$ in Eq. \ref{eq: OT-compact} is provided in the Appendix.




\noindent \textbf{Evaluation Metric:} In line with FG-SBIR research, we use Acc.@q accuracy, i.e. percentage of sketches having true matched photo in the top-q list. To evaluate performance on the partial setup, we explicitly mask local regions in scene and object sketch regions, described later in Sec. \ref{sec:evaluate-scene} and \ref{sec:evaluate-object} respectively. We repeat our evaluations $10$ times and report the average metrics. Since our goal is to evaluate \emph{robustness} to partial sketches, we train the network using the complete sketches but evaluate on \emph{masked partial sketch}.


\subsection{Competitors:}
\label{sec:competitors}
We compare against (i) existing state-of-the-art (SOTA) methods: \textbf{Triplet-SN} \cite{yu2016shoe} employs Sketch-A-Net \cite{sketch-a-net} backbone trained using triplet loss. \textbf{HOLEF} \cite{deep-spatial-semantic} extends \emph{Triplet-SN} by adding spatial attention along with higher order ranking loss. \textbf{On-the-fly} \cite{on-the-fly} aims to model partial sketches by employing reinforcement learning (RL) based fine-tuning for on-the-fly retrieval on partial object sketches. Since it requires vectored sketch data unavailable in existing scene sketch datasets \cite{sketchyscene, gao2020sketchyCOCO}, we compare with \emph{On-the-fly} on object sketches \cite{yu2016shoe} using vectored data. \cut{In addition, RL based training is unstable and requires longer training time.} As early retrieval is not our objective, we cite result at sketch completion point.  \textbf{SketchyScene} is the first work on scene-level FG-SBIR that adopts \emph{Triplet-SN} but replace the base network from Sketch-A-Net to InceptionV3 \cite{inceptionV3} along with an auxiliary cross-entropy loss to utilise the available object category information. \textbf{SceneSketcher} \cite{liu2020scenesketcher} is a recent work that use graph convolutional network to model the scene sketch layout information. (ii) Since local-level features are largely ignored by SOTAs, we design a few Baselines that computes the distance between localised sketch and photo features: \textbf{Local-Align} use a naive approach of computing cosine distance between a pair of local features at the same location of feature map from sketch and photo. However, sketches and photo do not follow strict region-wise alignment. \textbf{Local-MIL} on the other hand, utilise multiple instance learning (MIL) paradigm \cite{mil-sparse} where the minimum cosine distance pair between the set of localised sketch and photo is considered in loss computation. While it overcomes the limitation of misaligned localised features in \emph{Local-Align}, MIL is unstable since it leaves the remaining pairs of localised features under-constrained. \textbf{Local-Self-Atten} injects global context to local patches using self-attention mechanism \cite{self-atten} to aggregate contextual information followed by computing the cosine distance between localised feature maps as \emph{Local-Align}. (iii) {Naively computing a distance between localised features from sketch and photo fails to capture the geometry of the underlying localised feature space \cite{ot-vs-mmd}. Hence, we design two Variants of our proposed method: \textbf{Ours-Local-MMD} removes region-wise associativity from our proposed method and replace optimal transport with maximum mean discrepancy (MMD) to compare localised sketch and photo features that take into account the geometry of the underlying feature space. \textbf{Ours-Local-OT} replace MMD with the more accurate \cite{ot-vs-mmd} optimal transport to compute region-wise associativity between localised features.}

\setlength{\tabcolsep}{3pt}
\begin{table}[]
    \centering
    \footnotesize{
    \caption{Scene-level Fine-Grained SBIR on SketchyScene.}
    \vspace{-0.3cm}
    \begin{tabular}{clcccc}
        \toprule
         & {Method} &  & {\footnotesize{\makecell{Complete\\Sketch}}} & \makecell{$p_{mask}$\\0.3} & \makecell{$p_{mask}$\\0.5} \\ \hline
        \multirow{6}{*}{\rotatebox{90}{\textbf{SOTA}}} & \multirow{2}{*}{Triplet-SN \cite{yu2016shoe}} & Acc.@1 & 4.5 & $<$0.1 & $<$0.1 \\
         & & Acc.@10 & 26.7 & 6.9 $\pm$ 0.5 & 3.7 $\pm$ 0.7 \\ \cdashline{3-6}
         & \multirow{2}{*}{HOLEF \cite{deep-spatial-semantic}} & Acc.@1 & 5.3 & $<$0.1 & $<$0.1 \\
         &  & Acc.@10 & 29.5 & 7.5 $\pm$ 0.3 & 3.7 $\pm$ 0.7 \\ \cdashline{3-6}
         & \multirow{2}{*}{SketchyScene \cite{sketchyscene}} & Acc.@1 & 32.2 & 8.1 $\pm$ 0.7 & 4.5 $\pm$ 0.6 \\
         & & Acc.@10 & 69.3 & 24.7 $\pm$ 0.3 & 16.8 $\pm$ 1.3 \\ \hline
        \multirow{6}{*}{\rotatebox{90}{\textbf{Baselines}}} & \multirow{2}{*}{Local-Align} & Acc.@1 & 32.7 & 8.3 $\pm$ 0.4 & 4.9 $\pm$ 0.7 \\
         & & Acc.@10 & 70.1 & 31.6 $\pm$ 0.7 & 19.7 $\pm$ 1.2 \\ \cdashline{3-6}
         & \multirow{2}{*}{Local-MIL} & Acc.@1 & 33.4 & 9.7 $\pm$ 0.3 & 5.5 $\pm$ 0.6 \\
         & & Acc.@10 & 71.9 & 35.3 $\pm$ 0.5 & 21.2 $\pm$ 1.1 \\ \cdashline{3-6}
         & \multirow{2}{*}{Local-Self-Attn} & Acc.@1 & 33.6 & 9.7 $\pm$ 0.5 & 5.9 $\pm$ 0.6 \\
         & & Acc.@10 & 72.5 & 37.2 $\pm$ 0.8 & 21.7 $\pm$ 0.9 \\ \hline
        \multirow{4}{*}{\rotatebox{90}{\textbf{Variant}}} & \multirow{2}{*}{Ours-Local-OT} & Acc.@1 & 34.9 & 15.5 $\pm$ 0.3 & 8.1 $\pm$ 0.8 \\
         & & Acc.@10 & 74.5 & 45.3 $\pm$ 0.4 & 29.3 $\pm$ 0.7 \\ \cdashline{3-6}
         & \multirow{2}{*}{Ours-Local-MMD} & Acc.@1 & 34.4 & 14.5 $\pm$ 0.2 & 7.5 $\pm$ 0.7 \\
         & & Acc.@10 & 73.4 & 43.8 $\pm$ 0.5 & 25.9 $\pm$ 1.1 \\ \hline
         & \multirow{2}{*}{\textbf{Proposed Method}} & Acc.@1 & 35.7 & 20.6 $\pm$ 0.3 & 10.6 $\pm$ 0.9 \\
         & & Acc.@10 & 75.1 & 49.2 $\pm$ 1.1 & 29.9 $\pm$ 1.2 \\ \bottomrule
    \end{tabular}
    \label{tab:scene-sketchyscene}}
    \vspace{-0.1in}
\end{table}

\begin{figure}
    \centering
    \includegraphics[width=\linewidth]{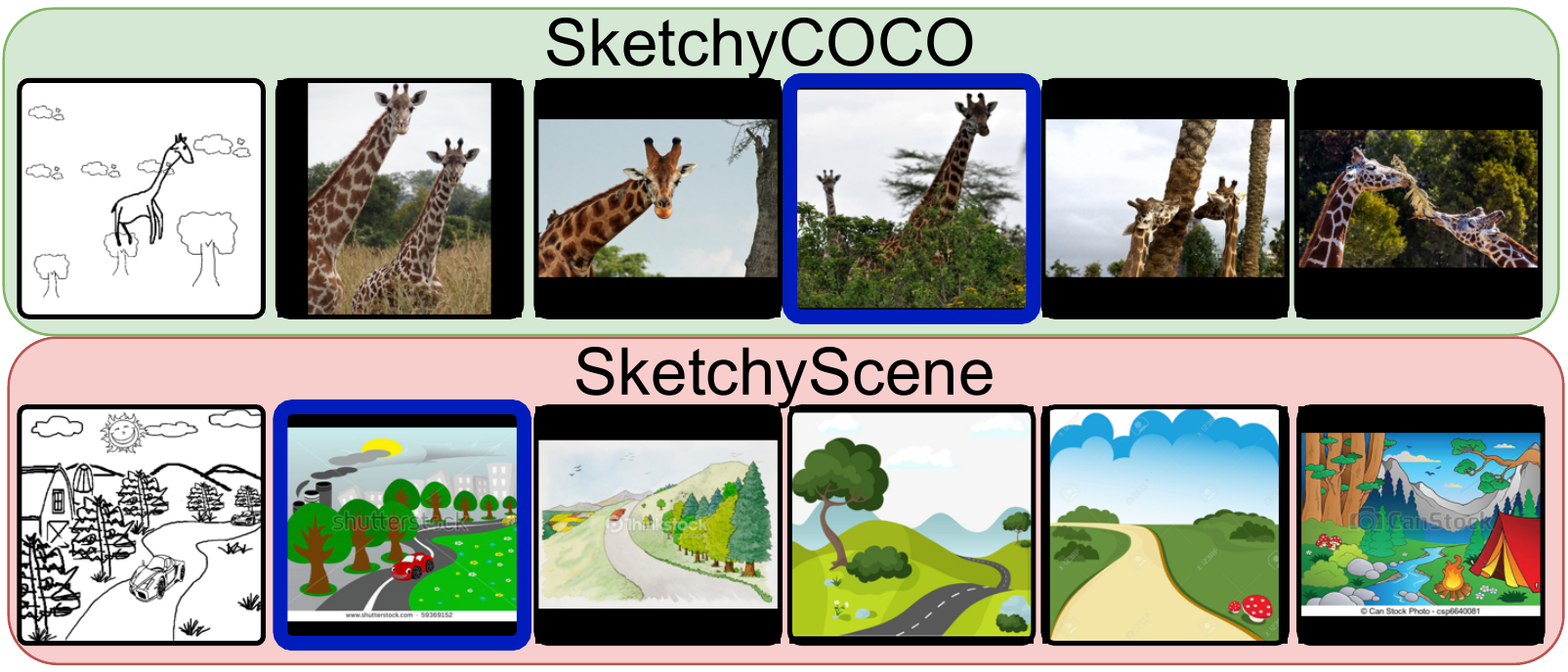}
    \vspace{-0.3in}
    \caption{Qualitative top-5 retrieved results on SketchyCOCO\cite{gao2020sketchyCOCO} and SketchyScene \cite{sketchyscene} using our proposed method. Blue denotes ground-truth photo.}
    \label{fig:retrieval}
    \vspace{-0.2in}
\end{figure}

\setlength{\tabcolsep}{3pt}
\begin{table}[]
    \centering  
    \footnotesize{
    \caption{Scene-level Fine-Grained SBIR on SketchyCOCO.}
    \vspace{-0.3cm}
    \begin{tabular}{clcccc}
        \toprule
         & {Method} & Acc.@q & {\footnotesize{\makecell{Complete\\Sketch}}} & \makecell{$p_{mask}$\\0.3} & \makecell{$p_{mask}$\\0.5} \\ \hline
        \multirow{8}{*}{\rotatebox{90}{\textbf{SOTA}}} & \multirow{2}{*}{Triplet-SN \cite{yu2016shoe}} & Acc.@1 & 6.2 & $<$0.1 & $<$0.1 \\
         & & Acc.@10 & 32.8 & 24.2 $\pm$ 0.9 & 18.5 $\pm$1.1 \\ \cdashline{3-6}
         & \multirow{2}{*}{HOLEF \cite{deep-spatial-semantic}} & Acc.@1 & 6.8 & $<$0.1 & $<$0.1 \\
         &  & Acc.@10 & 35.9 & 25.3 $\pm$ 0.7 & 19.3 $\pm$ 1.5 \\ \cdashline{3-6}
         & \multirow{2}{*}{SketchyScene \cite{sketchyscene}} & Acc.@1 & 27.6 & 19.7 $\pm$ 1.1 & 13.9 $\pm$ 1.4 \\
         & & Acc.@10 & 75.0 & 60.7 $\pm$ 1.2 & 48.0 $\pm$ 1.8 \\ \cdashline{3-6}
         & \multirow{2}{*}{SceneSketcher \cite{liu2020scenesketcher}} & Acc.@1 & 31.7 & 23.5 $\pm$ 1.5 & 17.2 $\pm$ 1.2 \\
         & & Acc.@10 & 86.1 & 70.7 $\pm$0.9 & 57.7 $\pm$ 1.3 \\ \hline
        \multirow{6}{*}{\rotatebox{90}{\textbf{Baselines}}} & \multirow{2}{*}{Local-Align} & Acc.@1 & 31.9 & 23.6$\pm$1.4 & 17.4$\pm$1.1 \\
         & & Acc.@10 & 86.6 & 70.7 $\pm$1.1 & 57.9$\pm$1.2 \\ \cdashline{3-6}
         & \multirow{2}{*}{Local-MIL} & Acc.@1 & 32.5 & 23.7$\pm$1.2 & 17.7$\pm$1.2 \\
         & & Acc.@10 & 87.8 & 71.1 $\pm$0.9 & 58.5$\pm$1.3 \\ \cdashline{3-6}
         & \multirow{2}{*}{Local-Self-Attn} & Acc.@1 & 33.1 & 23.9$\pm$1.1 & 18.1$\pm$1.4 \\
         & & Acc.@10 & 88.7 & 71.8 $\pm$ 0.7 & 59.1$\pm$1.1 \\ \hline
        \multirow{4}{*}{\rotatebox{90}{\textbf{Variant}}} & \multirow{2}{*}{Ours-Local-OT} & Acc.@1 & 34.3 & 24.7$\pm$1.2 & 18.7$\pm$1.5 \\
         & & Acc.@10 & 89.2 & 75.6 $\pm$ 1.1 & 65.7$\pm$1.2 \\ \cdashline{3-6}
         & \multirow{2}{*}{Ours-Local-MMD} & Acc.@1 & 33.9 & 24.2$\pm$0.9 & 18.5$\pm$1.3 \\
         & & Acc.@10 & 89.1 & 74.6$\pm$0.9 & 63.2$\pm$1.5 \\ \hline
         & \multirow{2}{*}{\textbf{Proposed Method}} & Acc.@1 & 34.5 & 25.1 $\pm$ 1.9 & 19.2 $\pm$ 1.4 \\
         & & Acc.@10 & 89.3 & 78.3 $\pm$ 1.6 & 69.3 $\pm$ 1.7 \\ \bottomrule
    \end{tabular}
    \label{tab:scene-sketchycoco}}
    \vspace{-0.1in}
\end{table}

\setlength{\tabcolsep}{3pt}
\begin{table}[]
    \centering
    \footnotesize{
    \caption{Object-level Fine-Grained SBIR on QMUL-Shoe-V2.}
    \vspace{-0.3cm}
    \begin{tabular}{clcccc}
        \toprule
         & {Method} & Acc.@q & {\footnotesize{\makecell{Complete\\Sketch}}} & \makecell{$p_{mask}$\\0.3} & \makecell{$p_{mask}$\\0.5} \\ \hline
        \multirow{6}{*}{\rotatebox{90}{\textbf{SOTA}}} & \multirow{2}{*}{Triplet-SN \cite{yu2016shoe}} & Acc.@1 & 28.7 & 22.3$\pm$0.4 & 9.7$\pm$0.9 \\
         & & Acc.@10 & 79.6 & 73.5$\pm$0.3 & 67.1$\pm$0.5 \\ \cdashline{3-6}
         & \multirow{2}{*}{HOLEF \cite{deep-spatial-semantic}} & Acc.@1 & 31.2 & 24.6$\pm$0.6 & 12.9$\pm$1.0 \\
         &  & Acc.@10 & 81.4 & 75.1$\pm$0.5 & 68.4$\pm$0.9 \\ \cdashline{3-6}
         & \multirow{2}{*}{On-the-fly \cite{on-the-fly}} & Acc.@1 & 34.1 & 29.5 $\pm$ 0.5 & 20.9 $\pm$ 0.9 \\
         & & Acc.@10 & 79.6 & 76.3$\pm$0.3 & 71.9$\pm$1.2 \\ \hline
        \multirow{6}{*}{\rotatebox{90}{\textbf{Baselines}}} & \multirow{2}{*}{Local-Align} & Acc.@1 & 33.5 & 25.7$\pm$0.2 & 14.9$\pm$0.7 \\
         & & Acc.@10 & 79.6 & 75.6$\pm$0.3 & 69.5$\pm$0.8 \\ \cdashline{3-6}
         & \multirow{2}{*}{Local-MIL} & Acc.@1 & 35.5 & 29.9$\pm$0.1 & 21.0$\pm$0.9 \\
         & & Acc.@10 & 80.6 & 79.1$\pm$0.5 & 71.3$\pm$1.1 \\ \cdashline{3-6}
         & \multirow{2}{*}{Local-Self-Attn} & Acc.@1 & 37.1 & 31.6$\pm$0.3 & 21.7$\pm$0.7 \\
         & & Acc.@10 & 81.4 & 79.5$\pm$0.1 & 71.5$\pm$0.5 \\ \hline
        \multirow{4}{*}{\rotatebox{90}{\textbf{Variant}}} & \multirow{2}{*}{Ours-Local-OT} & Acc.@1 & 39.7 & 34.7$\pm$0.3 & 25.7$\pm$1.0 \\
         & & Acc.@10 & 82.9 & 80.5$\pm$0.1 & 73.4$\pm$0.5 \\ \cdashline{3-6}
         & \multirow{2}{*}{Ours-Local-MMD} & Acc.@1 & 38.2 & 33.6$\pm$0.4 & 24.3$\pm$0.6 \\
         & & Acc.@10 & 82.5 & 79.7$\pm$0.2 & 73.3$\pm$0.5 \\ \hline
         & \multirow{2}{*}{\textbf{Proposed Method}} & Acc.@1 & 39.9 & 35.3$\pm$0.2 & 25.9$\pm$0.7 \\
         & & Acc.@10 & 82.9 & 80.9$\pm$0.1 & 73.4$\pm$0.7 \\ \bottomrule
    \end{tabular}
    \label{tab:object-qmul-shoe-v2}}
    \vspace{-0.25in}
\end{table}

\subsection{Evaluation on partial scene sketches}
\label{sec:evaluate-scene}
We perform a comparative study on scene sketches from SketchyScene \cite{sketchyscene} and SketchyCOCO \cite{gao2020sketchyCOCO} datasets. {Our experimental setup includes: {Complete Sketch} that evaluates on the original input scene sketch. \texttt{$p_{mask}=0.3$} use instance segmentation map of scene sketches available in both datasets to mask $30\%$ sketched objects in a scene.} Similarly, \texttt{$p_{mask}=0.5$} mask $50\%$ sketched objects respectively.

\noindent \textbf{Performance Analysis:} From Tab. \ref{tab:scene-sketchyscene} and Tab. \ref{tab:scene-sketchycoco} we make the following observations: (i) Performance of all SOTAs degrade significantly when increasing $p_{mask}$ from $0.3$ to $0.5$. This verifies our intuition that using a global feature vector is sub-optimal for the partial scene sketch setup. (ii) Our Baselines using local features are more resilient than SOTAs for partial scene sketches setup. It justifies the need for modelling localised features in partial scene sketches. However, abstract scene sketches and photos do not have a strict spatial alignment of localised regions assumed by \emph{Local-Align}. It explains the inferior performance of \emph{Local-Align} in comparison with other Baselines.
\emph{Local-MIL} considers only the minimum distance pair from a set of localised features in sketch and photo for loss computation but leaves the other pairs \emph{unconstrained}. This leads to instability during training which explains the inferior performance than \emph{Local-Self-Atten}. (iii) Both \emph{Ours-Local-MMD} and \emph{Ours-Local-OT} outperforms Baselines due to their ability to capture geometric information in the underlying localised feature space. Performance of \emph{Ours-Local-OT} is slightly superior to \emph{Ours-Local-MMD} due to the better accuracy of optimal transport in modelling the underlying geometry of localised feature space. Finally, our proposed method, equipped with optimal transport for region-wise associativity and weighted affinity matrix for scene structure consistency outperforms all competitors. Fig.~\ref{fig:retrieval} shows qualitative retrieval results on scene-sketch datasets.

\subsection{Evaluation on partial object sketches}
\label{sec:evaluate-object}
Most SOTAs in FG-SBIR were developed focusing on object-level sketches \cite{yu2016shoe, ha2018quickdraw, sketchy}. Hence for a fair comparison and to investigate the generalisation of our method to partial object-level sketches, we compare FG-SBIR using QMUL-Shoe-V2 \cite{yu2016shoe} dataset. {Our experimental setup comprise of: {Complete Sketch} that use the original object sketch. For $p_{mask}=0.3$, $p_{mask}=0.5$ we mask $30\%$ and $50\%$ of the strokes respectively. We derive strokes using the available point coordinate and pen state information.}

\noindent \textbf{Performance Analysis:}
From Tab. \ref{tab:object-qmul-shoe-v2}, we observe: (i) With the exception of \emph{On-the-fly}, performance of all SOTAs degrade for partial object sketch, as more strokes are masked from $p_{mask}=0.3$ to $p_{mask}=0.5$.  High performance of \emph{On-the-fly} among SOTAs is due to its reinforcement learning based fine-tuning that takes into account the complete episode of progressive complete sketch from $p_{mask}=0.5$ to {Complete Sketch} before updating the weights. This provides a more principled and practically reliable approach to modelling partial object sketches. However, its RL based unstable training leads to inferior performance than our Baselines. (ii) The performance gap between SOTAs and Baselines narrows down from that observed for scene sketches. In particular, \emph{On-the-fly} using global average pooling but explicitly trained on episodes of partial sketches outperforms \emph{Local-Align} trained on {Complete Sketches} and naively computes the distance between a pair of local sketch and photo features at the same location. This suggests that for object-level sketch, a carefully trained global feature can surpass naively training localised features. (iii) A lower performance gap between \emph{Ours-Local-OT} and the proposed method for object sketches when compared to that of scene sketch show the effect of weighted affinity matrix dilutes for the simpler scenario of object sketches.

\subsection{Ablation}

\noindent \textbf{Significance of Adjacency Matrix:} 
We quantitatively evaluate the significance of adjacency matrix in Tab. \ref{tab:ablation} where including weighted adjacency matrix (W-Adj) improves scene-level FG-SBIR Acc.@1 performance by 1.0\% and 1.4\% over \emph{SketchyScene} in complete and partial sketch (\texttt{$p_{mask}=0.3$}) respectively.

\begin{figure}
    \centering
    \includegraphics[width=\linewidth]{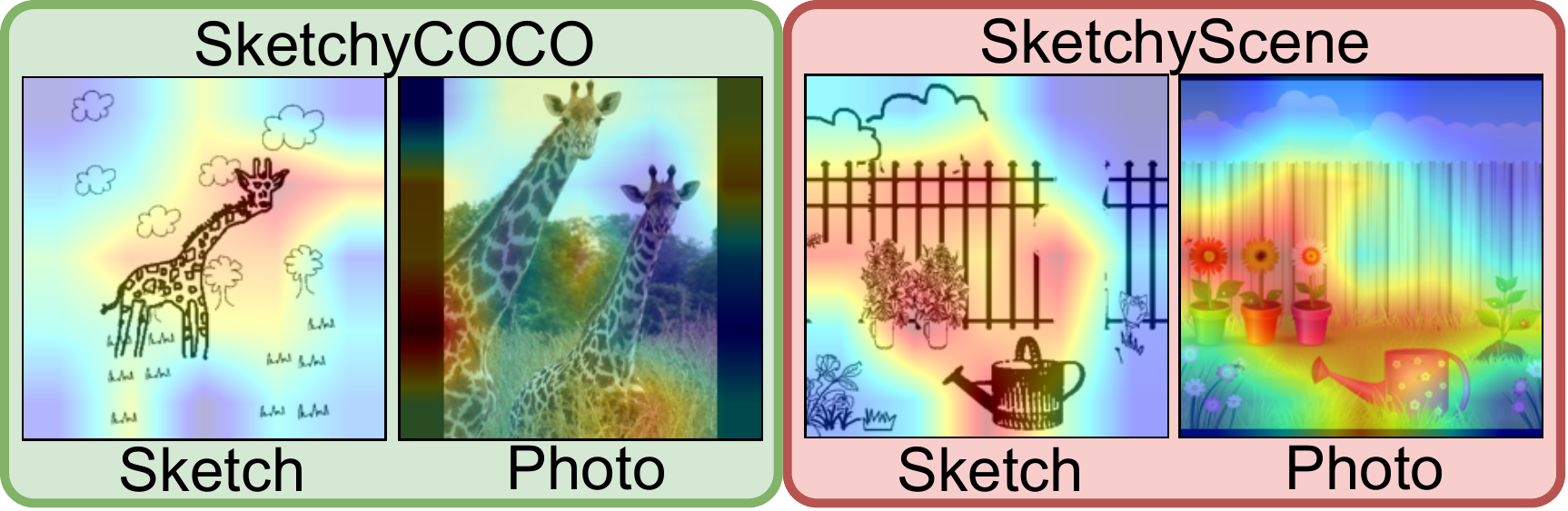}
    \vspace{-0.3in}
    \caption{{Illustrating region-wise associativity between ``partial" scene sketch and photo from SketchyCOCO \cite{gao2020sketchyCOCO} and SketchyScene \cite{sketchyscene}. For visualisation, we select the maximum value of region-wise associativity $(x_{i,j})$ from a local sketch region $(u_i)$ to all photo regions $\{v_j\}_{j=1}^{n}$, and vice-versa.}}
    \label{fig:attention-maps}
    \vspace{-0.2in}
\end{figure}

\noindent \textbf{Significance of weighting factor $\omega_{i,j}^{u,v}$:}
Partial sketches, by definition, contain empty local regions. Naively computing the adjacency matrix (Eq. \ref{eq:adjacency}) would unnecessarily penalise and confuse the model. 
The weighting factor $\omega_{i,j}^{u,v}$ makes the model \emph{aware} of empty regions by comparing $(u_i, u_j)$ across modality with $(v_i, v_j)$ in Eq. \ref{dg_eqn}. From Tab. \ref{tab:ablation}, adding naive adjacency matrix (N-Adj) enhance accuracy over \emph{SketchyScene} by 0.7\%, whereas a weighted adjacency matrix (W-Adj) gives 1.0\% improvement. Fig.~\ref{fig:attention-maps} highlights the corresponding sketch ($u_i$) with photo ($v_j$) regions.

\setlength{\tabcolsep}{5pt}
\begin{table}[]
    \centering
    \small{
    \caption{Ablative study on SketchyScene: \emph{(w/o)-Region-wise Associativity} (RwA), using Naive Adjacency matrix (N-Adj), Weighted Adjacency (W-Adj) from Eq. \ref{dg_eqn}.}
    \vspace{-0.3cm}
    \begin{tabular}{cccccc}
        \toprule
        \makecell{RwA} & \makecell{N-Adj} & \makecell{W-Adj} & \makecell{Complete\\Sketch} & \makecell{\texttt{$p_{mask}$}\\0.3} & \makecell{Test\\time} \\\hline
        \xmark & \xmark & \xmark & {32.2} & {8.1} & {11.9ms} \\
        \cmark & \xmark & \xmark & {34.9} & {15.5}  & {12.4ms} \\
        \xmark & \cmark & -- & {32.9} & {8.9}  & {12.0ms} \\
        \xmark & -- & \cmark & {33.2} & {9.5} & {12.0ms} \\\hline
        \cmark & -- & \cmark & {35.7} & {20.6} & {12.4ms} \\\bottomrule
    \end{tabular}
    \label{tab:ablation}}
    \vspace{-0.1cm}
\end{table}

\noindent \textbf{Computation Time:} From Tab. \ref{tab:ablation}, we observe that including region-wise associativity (RwA) increases computation time by 0.5ms per iteration. Although insignificant compared to existing SOTA \emph{Triplet-SN}, this added computation dramatically improves accuracy on complete and partial sketches. This slight increase in time is due to convex optimisation solvers is expected to diminish with faster and efficient solvers in the near future.

\section{Conclusion}
In this paper, we scrutinise an important observation plaguing scene-level sketch research -- that a significant portion of scene sketches are ``partial". We analyse how this partialness happens on two fronts: (i) significant empty (white) regions -- locally partial, (ii) scene sketch does not necessarily contain all objects in the corresponding photo -- holistically partial. For this, we propose a solution using a simple set-based approach using optimal transport to overcome local partialnes, and weighted cross-modal comparison of intra-modal adjacency matrices to address holistic partialness. Empirical evidence shows remarkable performance as a direct result of tackling this partial problem. 

{\small
\bibliographystyle{ieee_fullname}
\bibliography{main}
}

\cleardoublepage

\title{\vspace{-0.5cm}Partially Does It: Towards Scene-Level FG-SBIR with Partial Input. \\ Supplemental material\vspace{-0.5cm}}

\lstset{
    basicstyle=\ttfamily,
    keywordstyle=\color{blue},
    stringstyle=\color{DarkMagenta},
    commentstyle=\color{DarkGreen},
    morecomment=[l]{\%}
}

\author{Pinaki Nath Chowdhury\textsuperscript{1,2} \hspace{.2cm} Ayan Kumar Bhunia\textsuperscript{1} \hspace{.2cm} Viswanatha Reddy Gajjala$^{*}$ \\ \hspace{.2cm} Aneeshan Sain\textsuperscript{1,2} \hspace{.2cm} Tao Xiang\textsuperscript{1,2} \hspace{.2cm} Yi-Zhe Song\textsuperscript{1,2} \\
\textsuperscript{1}SketchX, CVSSP, University of Surrey, United Kingdom.  \\
\textsuperscript{2}iFlyTek-Surrey Joint Research Centre on Artificial Intelligence.\\
{\tt\small \{p.chowdhury, a.bhunia, a.sain, t.xiang, y.song\}@surrey.ac.uk}\vspace{-0.3cm}
\vspace{-0.5cm}
}

\maketitle

\renewcommand{\thesubsection}{\Alph{subsection}}
\setcounter{figure}{8}
\setcounter{table}{7}

\definecolor{commentcolor}{RGB}{110,154,155}   
\newcommand{\PyComment}[1]{\ttfamily\textcolor{commentcolor}{\# #1}}  
\newcommand{\PyCode}[1]{\ttfamily\textcolor{black}{#1}} 

\subsection{PyTorch-like pseudo-code to solve the linear programming problem using QPTH .}

\setlength{\textfloatsep}{0pt}
\begin{algorithm}[h]
\SetAlgoLined
 \footnotesize
    \PyCode{import torch} \\
    \PyComment{A differentiable QP solver for PyTorch} \\
    \PyCode{from qpth.qp import QPFunction} \\
    \PyCode{} \\
    \PyCode{def compute\_flow(u, v):} \\
    \Indp   
        \PyComment{u: Tensor of shape [nbatch,c,m]} \\
        \PyComment{v: Tensor of shape [nbatch,c,n]} \\
        \PyCode{nbatch, \_, m = u.shape} \\
        \PyCode{n = v.shape[2]} \\
        
        \PyCode{} \\
        \PyComment{Objective Function in Eq. 4} \\
        \PyCode{Q = 1e-3 * torch.eye(m*n).float()} \\
        \PyCode{Q = Q.unsqueeze(0).repeat(nbatch,1,1)} \\
        \PyCode{p = torch.bmm(u.permute(0,2,1), v)} \\
        \PyCode{p = p.view(nbatch, m*n)} \\
        
        \PyCode{} \\
        \PyComment{Inequality Constraint $x_{i,j}\geq0$} \\
        \PyCode{G = -torch.eye(m*n).float()} \\
        \PyCode{G = G.unsqueeze(0).repeat(nbatch,1,1)} \\
        \PyCode{h = torch.zeros(nbatch, m*n)} \\
        
        \PyCode{} \\
        \PyComment{Equality Constraint in Eq. 5} \\
        \PyCode{A = torch.zeros(nbatch, m+n, m*n)} \\
        \PyCode{for i in range(m):} \\
        \Indp
            \PyCode{A[:, i, n*i:n*(i+1)] = 1} \\
        \Indm 
        \PyCode{for j in range(n):} \\
        \Indp
            \PyCode{A[:, m+j, j::n] = 1} \\ 
        \Indm
        \PyCode{s = get\_weights(u, v)} \PyComment{(nbatch, m)} \\
        \PyCode{d = get\_weights(v, u)} \PyComment{(nbatch, n)} \\
        \PyCode{b = torch.cat([s, d], dim=1)} \\
        
        \PyCode{} \\
        \PyComment{flow $\mathcal{\hat{X}}$ shape:(nbatch,m,n) in Eq. 7} \\
        \PyCode{flow = QPFunction()(Q,p,G,h,A,b)} \\
        \PyCode{return flow.view(nbatch, m, n)} \\
    \Indm
    \PyCode{} \\
    \PyComment{Utility function for Eq. 5} \\
    \PyCode{def get\_weights(a, b):} \\
    \Indp
        \PyCode{node = a.shape[2]} \\
        \PyCode{w = a*b.sum(dim=2).repeat(1,1,node)} \\
        \PyCode{w = torch.relu(w.sum(dim=1)) + 1e-3} \\
        \PyCode{return w} \\

\caption{{PyTorch code to compute flow $\mathcal{\hat{X}}$}}
\label{alg:flow}
\end{algorithm}

\subsection{Additional Discussion}






\subsubsection{Why our proposed method outperform SceneSketcher, a method that uses bounding box annotation for scene graph matching?}
 Performance of graph based methods depend significantly on (1) graph construction step \cite{copt}, and (2) graph matching loss used for a downstream task \cite{graph-cm-OT}. This hints at the bottleneck of graph based approaches -- a sub-optimal graph, that is often constructed based on some heuristics (e.g., computing cosine distance of selected foreground regions \cite{liu2020scenesketcher}), might lead to sub-optimal performance. The graph matching metric used in SceneSketcher \cite{liu2020scenesketcher} has a remarkable similarity to that of Multiple Instance Learning \cite{mil-sparse}, that computes a loss between the most similar pairs, but leaves the other pairs unconstrained. While one could adapt SceneSketcher using Gromov-Wasserstein distance \cite{gromov-wasserstein}, in this work, we advocate for a graph-free approach that do not need expensive bounding box annotations.

\subsubsection{Why not train on partial sketches?} 

While training on partial sketches can \emph{artificially} inflate retrieval performance during evaluation, the objective of this paper is to study robustness of scene-level FG-SBIR methods for partial or incomplete sketches -- especially for scenes where the problem is most relevant, as shown in our pilot study in Sec. 1. In addition, {the strategy used to mask local sketch regions can have significant effect on performance of the model \cite{on-the-fly}}. Hence, instead of relying on tricks based on heuristics to improve performance, our objective is to propose a distance function which is implicitly robust to partial sketches with a well studied theoretical background that popular in the research community.


\subsubsection{Understanding the dilemma between \emph{fast}- and \emph{slow}-retrieval:}

There can be two major approaches to fine-grained image retrieval, a 
\emph{fast}, and a \emph{slow} retrieval: (i) In \emph{fast} retrieval, photos and sketches are embedded independently into a joint embedding space and then their similarities are compared. We pre-compute the feature vectors for each photo in the gallery independently, prior to having access of any query sketch. During inference, a single pass through the encoder is performed to embed the input sketch query to the joint sketch-photo embedding space. The resulting feature vector is then matched to its semantically similar photo using some distance function (usually euclidean or cosine distance \cite{yu2016shoe, deep-spatial-semantic}). Given $n$ photos in the gallery set, one would spend $\mathcal{O}(1)$ forward pass through encoder network. (ii) On contrary, \emph{slow}-retrieval models trade off compute time for accuracy gains. They explore the interactions between photos and query sketch \emph{before} calculating similarities in the joint sketch-photo embedding space. Existing methods like Wang \etal \cite{camp}, propose to adaptively control the information flow for message passing across modalities. However, a key limitation to adaptively updating sketch and photo features is that we can only compute \emph{paired}-feature embedding that jointly represents similarity of a sketch-photo pair. Considering $n$ photos in our gallery during inference, we have to compute the paired embedding of a given query sketch with each photo that needs $\mathcal{O}(n)$ forward pass through the network. For practical applications where $n$ can be millions of photos, $\mathcal{O}(n)$ forward pass through a heavy neural network is intractable.

We propose a \emph{mid-ground} between \emph{fast}- and \emph{slow}- retrieval. Instead of computing paired sketch-photo embedding, we propose to independently compute local-level feature maps for each sketch and photo. Our novel distance function, then \emph{adaptively} computes region-wise features from sketch and photo using region-wise associativity that gives greater weightage to semantically similar local patches. Since we independently compute local-level features, during inference, our approach needs $\mathcal{O}(1)$ forward pass through a neural network. Although our simple trick can result in competitive performance to \emph{slow}-retrieval models, storing local-level features increase the space complexity. Effective approaches in annealing space complexity could be an interesting direction of future research.


\end{document}